\documentclass{clv3}

\usepackage{hyperref}
\usepackage{xcolor}
\usepackage{graphicx}
\usepackage{tabularx} 

\definecolor{darkblue}{rgb}{0, 0, 0.5}
\hypersetup{colorlinks=true,citecolor=darkblue, linkcolor=darkblue, urlcolor=darkblue}

\begin{document}

\title{Combining pre-trained language models and structured knowledge}

\author{Pedro Colon-Hernandez\thanks{75 Amherst St, Cambridge, MA, 02139. E-mail: pe25171@mit.edu.}}
\affil{MIT Media Lab}

\author{Catherine Havasi}
\affil{Dalang Health}

\author{Jason Alonso}
\affil{Dalang Health}

\author{Matthew Huggins}
\affil{MIT Media Lab}

\author{Cynthia Breazeal}
\affil{MIT Media Lab}
\maketitle

\begin{abstract}
In recent years, transformer-based language models have achieved state of the art performance in various NLP benchmarks.  These models are able to extract mostly distributional information  with some semantics from unstructured text, however it has proven challenging to integrate structured information, such as knowledge graphs into these models.  We examine a variety of approaches to integrate structured knowledge into current language models and determine challenges, and possible opportunities to leverage both structured and unstructured information sources. From our survey, we find that there are still opportunities at exploiting adapter-based injections and that it may be possible to further combine various of the explored approaches into one system.   
\end{abstract}

\section{Introduction}

Recent developments in Language Modeling (LM) techniques have greatly improved the performance of systems in a wide range of Natural Language Processing (NLP) tasks.  Many of the current state of the art systems are based on variations to the transformer \cite{vaswani2017attention} architecture. The transformer architecture, along with modifications such as the Transformer XL \cite{dai2019transformer} and various training regimes such as the Masked Language Modeling (MLM) used in BERT \cite{devlin2018bert} or the Permutation Language Modeling (PLM) used in XLNet\cite{yang2019xlnet}, uses an attention based mechanism to model long range dependencies between text.  This modeling encodes syntactic knowledge, and to a certain extent some semantic knowledge contained in unstructured texts.  

There has been interest in being able to understand what kinds of knowledge are encoded in these models' weights.  Hewitt et al. \cite{hewitt2019structural} devise a system that generates a distance metric between embeddings for words in language models such as BERT.  They show that there is some evidence that there are syntax trees embedded in the transformer language models and this could explain the performance of these models in some tasks that utilize syntactic elements of text. 

Petroni et al. \cite{petroni2019language} build a system (LAMA) to gauge what kinds of knowledge are encoded in these weights. They discover that language models embed some facts and relationships in their weights during pre-training. This in turn can help explain the performance of these models in semantic tasks.  However, these transformer-based language models have some tendency to hallucinate knowledge (whether through bias or incorrect knowledge in the training data).  This also means that some of the semantic knowledge they incorporate is not rigidly enforced or utilized effectively. 

Avenues of research have begun to open up on how to prevent this hallucination and how to inject additional knowledge from external sources into the transformer-based language models.  One  promising avenue is through the integration of knowledge graphs such as Freebase\cite{bollacker2008freebase}, WordNet\cite{miller1998wordnet}, ConceptNet\cite{speer2017conceptnet}, and ATOMIC\cite{sap2019atomic}. 

A knowledge graph (used somewhat interchangeably with knowledge base although they are different concepts) is defined as ``a graph of data
intended to accumulate and convey knowledge of the real world, whose nodes represent entities of
interest and whose edges represent relations between these entities" \cite{hogan2020knowledge}.  Formally, a knowledge graph is a set of triples that represents nodes and edges between these nodes.  Let us define a set of vertices (which we will refer to as concepts) as $V$, a set of edges as $E$ (which we will refer to as assertions as per Speer and Havasi\cite{speer2012representing}), and a set of labels $L$ (which we will refer to as relations). A knowledge graph is a tuple $G:=(V,E,L)$\footnote{We use the formal definitions found in Appendix B of \cite{hogan2020knowledge}}. The set of edges ($E$) or assertions is composed of triples $E \subseteq V\times L\times V$ which are seen as a subject (a concept), a relation (a label), and object (another concept) respectively (e.g. $(subject,relation,object)$).  These edges in some cases can have weights to represent the strength of the assertion. Broadly speaking, knowledge graphs (KGs) are a collection of tuples that represent things that should be true within the knowledge of the world that we are representing.  An example assertion is ``a dog is an animal" and its representation as a tuple would be: (dog,isA,animal).  

Ideally, we would want to ``\textit{inject}" this structured collection of confident information (i.e. knowledge graph) into that of the high-coverage, contextual information found in language models. This injection would permit the model to incorporate some of the information found in the KG to improve its performance in inference tasks. 

\begin{figure*}[ht!]
  \centering
  \includegraphics[width=0.9\textwidth,height=5cm]{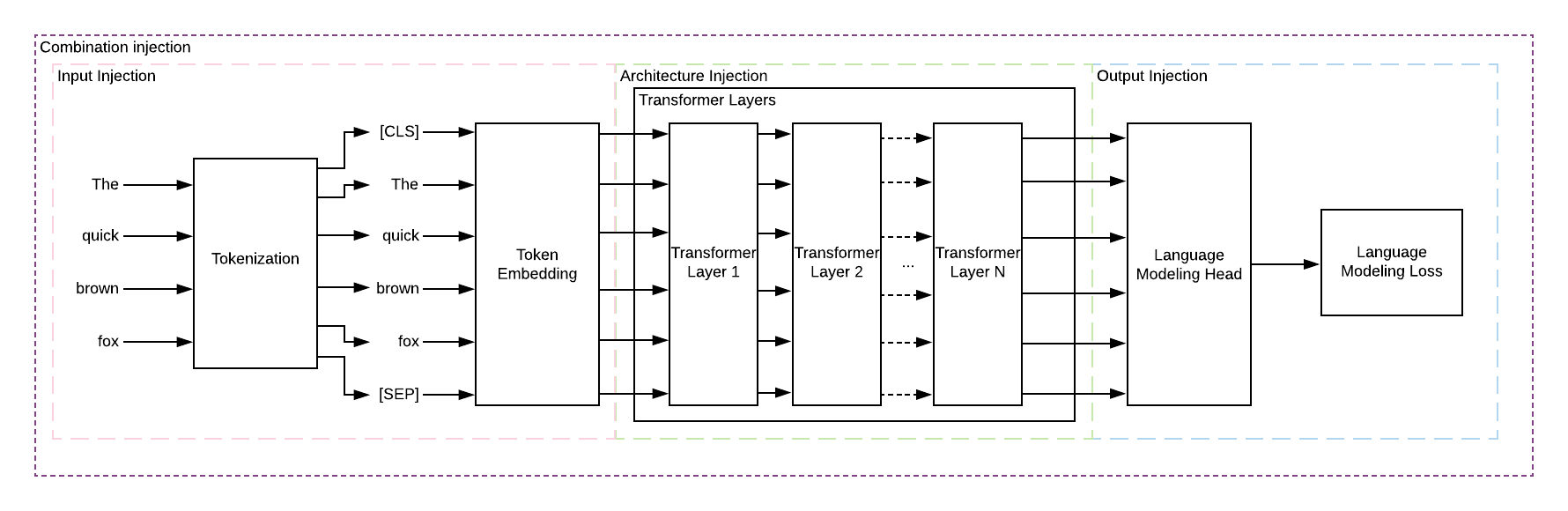}
  \caption{Visualization of boundaries of the different categories of knowledge injections. \textit{Combination} injections involve combinations of the three categories. }
  \label{fig:injectiontypes}
\end{figure*}

There are currently various approaches that try to achieve this injection. The approaches in general take either one or combinations of three forms: \textit{input} focused injections, \textit{architecture} focused injections, and \textit{output} focused injections.  We define an \textit{input} focused injection as any technique that modifies the data pre-processing or the pre-transformer layer inputs that the base model uses(i.e. injecting knowledge graph triples into the training data to pre-train/fine tune on them or combining entity embeddings into the static word embeddings that the models have).  We define \textit{architecture} focused injections as techniques that alter a base model's transformer layers (i.e. adding additional layers that inject in some representation).  Lastly, we define an \textit{output} focused injection as any techniques that either modify the output of the base models or that modify/add custom loss functions. In addition to these three basic types, there are approaches that utilize combinations of these (i.e. a system that uses both input and output injections), which we call \textit{combination} injections. Figure \ref{fig:injectiontypes} gives an abstract visualization of the types of injections that we describe.  

To be consistent throughout the type of injections, we will now give some definitions and nomenclature.  Let us define a sequence of words (unstructured text) as  $S$.  Typically in a transformer-based model, this sequence of words is converted to a sequence of tokens that is then converted into some initial context-independent embeddings. To a word sequence we can apply a tokenization technique $\mathcal{T}$ to convert the word sequence into a token sequence $T$.  This can be seen as $\mathcal{T} (S) = T$. This sequence $T$ is used as a lookup in an embedding layer $\mathcal{E}$ to produce context independent token vector embeddings: $\mathcal{E}(T) = E$. These are then passed sequentially through various contextualization layers (i.e. transformers) which we define as the set  $\mathcal{H}$, $\mathcal{H}=(H_1,...,H_n)$.  The successive application of these ultimately produces a sequence of contextual embeddings $C$: $C=H_n(H_{n-1}(...H_1(E)))$. We additionally define $\mathcal{G}$ as graph embeddings of a knowledge graph $G$ that are the result of some embedding function $\mathcal{E}_g$: $\mathcal{G} = \mathcal{E}_g(G)$.  This final sequence is run through a final layer $L_{LM}$ that is used to calculate the language modeling loss function $\mathcal{L}$ that is optimized through back-propagation.  The notation that we utilize is intentionally vague on the definition of the  functions, in order for us to fit the different works that we survey.    

In the following sections we will look at attempts of injecting knowledge graph information that fall into the aforementioned categories. Additionally, we will highlight relevant benefits in these approaches.  We conclude with possible opportunities and future directions for injecting structured knowledge into language models.

\section{Input Focused Injections}
\label{sec:inputfocusedinjections}
In this section we will describe knowledge injections whose techniques center around modifying either the structure of the input or the data that is selected to be fed into the base transformer models. A common approach to inject information from a knowledge graph is by converting its assertions into a set of words (possibly including separator tokens)  and pre-training or fine-tuning a language model with these inputs. We discuss two particular papers that focus on structuring the input in different ways as to capture the semantic information from triples found in a KG. These approaches start from a pre-trained model, and fine-tune on their knowledge infusing datasets. A summary of these approaches can be found in table \ref{tab:inputinjections}. 

Input focused injections can be seen any technique whose output is a modified $E$, hereby known as $E'$.  This modification can be achieved either by modifying $S$,$\mathcal{T}$, $T$, $\mathcal{E}$,or directly $E$. (i.e. the word sequence, the token sequence, the tokenization function, the context-less embedding function, or the actual context-less embeddings).  The hope of \textit{input} focused injections is that the knowledge in $E'$ will be distributed and contextualized through $\mathcal{H}$ as the language models are trained. 

\subsection{Align, Mask, Select (AMS) \cite{ye2019align}}
AMS is an approach  in which a question answering dataset is created, whose questions and possible answers are generated by aligning a knowledge graph (in this particular case ConceptNet) with plain text. A BERT model is trained on this dataset to inject it with the knowledge. 

Taking an example from their work, the ConceptNet triple \textit{(population, AtLocation, city)} is aligned  with a sentence from the English Wikipedia (i.e. ``The largest \textbf{city} by \textbf{population} is Birmingham, which
has long been the most industrialized city.") that contains both concepts in the triple. They then proceed to mask out one of the concepts with a special token $([QS])$ and produce 4 plausible concepts as answers to the masking task by looking at the neighbors in ConceptNet that have the same masked token and relationship.   Lastly, they concatenate the generated question with the plausible answers and run it through a BERT model tailored for question answering (QA) (following the same approach as the architecture and loss for the SWAG task in the original BERT).  At the output, they run the classification token $([CLS])$ through a softmax classifier to determine if the selected concept is the correct one or not.  

The authors note that the work is sensitive to what it has seen in the pre-training because when asked a question that needs to disambiguate a pronoun it tries to match what it has seen the most in the training data.  This may mean that the generalization of the structured knowledge (here commonsense information) or the understanding of the it is overshadowed by the distributional information that it is learning, however more testing would need to be done to verify this.  Overall some highlights of the work are:
\begin{itemize}
  \item Automated pre-training approach which constructs a QA dataset aligned to a KG
  \item Utilization of graph-based confounders in generated dataset entries
\end{itemize}

\subsection{COMmonsEnse Transformers (COMET) \cite{bosselut2019comet}}
COMET is a GPT\cite{radford2018improving} based system which is trained on triples from KGs (ConceptNet and Atomic) to learn to predict the object of the triple (the triples being defined as (subject, relation, object)).  The triples are fed as a concatenated sequence of words into the model (i.e. the words for the subject, the relationship, and the object) along with some separators. 

The authors initialize the GPT model to the final weights in the training from Radford et al.\cite{radford2018improving} and proceed to train it to predict the words that belong to the object in the triple.  A very interesting part of this work is that it is capable directly of performing knowledge graph completion for nodes and relations that may not have been seen during training, in the form of sentences. 

Some plausible shortcomings of this work is that you are still having the model extract the semantic information from the distributional one and possibly suffering from the same bias as AMS.  In addition to this, by training on the text version of these triples, it may be the case that we lose some of the syntax that the model learns due to awkwardly formatted inputs (i.e. ``cat located at housing" rather than ``a cat is located at a house"), however further testing of these two needs to be performed. 

There is some relevant derivative work for COMET by Bosselut et al.\cite{bosselut2019dynamic} which looks into how effective COMET is at building KGs on the fly given a certain context, a question, and a proposed answer. They combine the context with a relation from ATOMIC and feed it into COMET to represent reasoning hops. They do this for multiple relations and keep redoing this with the generated outputs to represent a reasoning chain for which they can derive a probability. They use this in a zero-shot evaluation of a question-answering system and find that it is effective. 
Overall some highlights of COMET are:
\begin{itemize}
    \item Generative language model that can provide natural language representations of triples
    \item Useful model for zero-shot KG completion
    \item Simple pre-processing of triples for training
\end{itemize}

\begin{table*}[ht!]
    \centering
    \small
    \begin{tabularx}{\textwidth}{|m{1cm}|X|X|}
\hline
  \textbf{Model} & \textbf{Summary of Injection}& \textbf{Example of Injection}\\\hline
  Align, Mask, Select&	Aligns a knowledge base with textual sentences, masks entities in the sentences, and selects alternatives with confounders to create a QA dataset&	\textbf{KG Assertion}: (population, AtLocation, city)\newline
 \textbf{Model Input}: The largest [QW] by population is Birming- ham, which has long been the most industrialized city?  city, Michigan, Petrie dish, area with people inhabiting, country\\\hline
COMET&Ingests a formatted sentence version of a triple from ConceptNet and Atomic&\textbf{KG Assertion}: 
(PersonX goes to the mall, xIntent, to buy clothes)\newline
\textbf{Model input}:
PersonX goes to the mall [MASK] $\langle$ xIntent $\rangle$ 
to buy clothes\\\hline

  \end{tabularx}
    \caption{Input Injection System Comparisons}
    \label{tab:inputinjections}
\end{table*}

\section{Architecture Injections}
In this section we describe approaches that focus on architectural changes to language models. This involves either adding additional layers that integrate knowledge in some way with the contextual representations or modifying existing layers to manipulate things such as attention mechanisms. We discuss two approaches within this category that fall under layer modifications.  These approaches utilize adapter-like mechanisms to be able to inject information into the models.  A summary of these approaches can be found in table \ref{tab:archinjections}.

\subsection{KnowBERT\cite{peters2019knowledge}}
KnowBERT modifies BERT's architecture by integrating some layers that they call the Knowledge Attention and Recontextualization (KAR).  These layers take graph entity embeddings, that are based on Tucker Tensor Decompositions for KG completion \cite{balavzevic2019tucker}, and run them through an attention mechanism to generate entity span embeddings.  These span embeddings are then added to the regular BERT contextual representations.  The summed representations are then uncompressed and passed on to the next layer in a regular BERT. Once the KAR entity linker has been trained, the rest of the BERT model is unfrozen and is trained in the pre-training. These KAR layers are trained for every KG that is to be injected, in this work they use data from Wikipedia and Wordnet.  

An interesting observation is that the injection happens in the later layers, which means that the contextual representation up to that point may be unaltered by the injected knowledge.  This is done to stabilize the training, but could present an opportunity to inject knowledge in earlier levels. Additionally, the way the system is trained, the entity linking is first trained, and then the whole system is unfrozen to incorporate the additional knowledge into BERT.  This strategy could lead to the catastrophic forgetting\cite{kirkpatrick2017overcoming} problem where the knowledge from the underlying BERT model or the additional structured injection may be forgotten or ignored.    

This technique falls into a broader category of what is called Adapters\cite{houlsby2019parameter}.  Adapters are layers that are added into a language model and are subsequently fine tuned to a specific task.  The interesting aspect of adapters is that they add a minimal amount of additional parameters, and freeze the original model weights.  The added parameters are also initialized to produce a close to identity output. It is worth noting that the KnowBERT is not explicitly an Adapter technique as the model is unfrozen during training. Some highlights of KnowBERT are the following:
\begin{itemize}
    \item Fusion of contextual and graph representation of entities
    \item Attention enhanced entity spanned knowledge infusion
    \item Permits the injection of multiple KGs in varying levels of the model
\end{itemize}

\subsection{Common  sense  or  world  knowledge?  investigating adapter-based knowledge injection  into  pre-trained  transformers\cite{lauscher2020common} }
This work explores what kinds of knowledge are infused by fine tuning an adapter equipped version of BERT on ConceptNet. They generate and test models trained on sentences from the Open Mind Common Sense (OMCS)\cite{singh2002open} corpus and from walks in the ConceptNet graph.  They note that with simple adapters and as little as 25k/100k update steps on their training sentences, they are able greatly improve the encoded ``World Knowledge" (another name for the knowledge found in ConceptNet).  

However, it is worth noting that the information is presented as sentences to which the adapters are fine tuned. This may mean that the model may have similar possible shortcomings such as with the approaches that are input-focused (model may rely more on the distributional rather than the semantic information), however testing needs to be performed to confirm this.     Overall some highlights of this work are the following:
\begin{itemize}
    \item Adapter based approach which fine-tunes a minimal amount of parameters
    \item Shows that a relatively small amount of additional iterations can inject the knowledge in the adapters
    \item Show that adapters, trained on KGs, do indeed boost the semantic performance of transformer-based models
\end{itemize}

\begin{table*}[ht!]
    \centering
    \small
    \begin{tabularx}{\textwidth}{|X|m{1.75cm}|m{1.75cm}|X|}
\hline
  \textbf{Model} & \textbf{Injected in Pre-Training}& \textbf{Injected in Fine-Tuning}& \textbf{Summary of Injection}\\\hline
  KnowBERT&Yes&Yes&Sandwich Adapter-like layers which sum contextual representation of layer with graph representation of entities and distributes it in an entity span\\\hline
  Common  sense  or  world  knowledge?[...]&No&Yes&Use sandwich adapters to fine tune on a KG \\\hline

  \end{tabularx}
    \caption{Architecture Injection System Comparisons}
    \label{tab:archinjections}
\end{table*}

\section{Output Injections}
In this section we describe approaches that focus on changing either the output structure or the losses that were used in the base model in some way to incorporate knowledge. Only one model falls strictly under this category, the model injects entity embeddings into the output of a BERT model.    

\subsection{SemBERT\cite{zhang2019semantics}}

SemBERT uses a subsystem that generates embedding representations of the output of a semantic role labeling\cite{marquez2008semantic} system. They then concatenate this representation with the output of the contextualized representation from BERT to help incorporate relational knowledge. The approach, although clever, may fall short in that although it gives a representation for the roles, it leaves the model to figure out the exact relationship that the roles are performing, however testing would need to be performed to check this. Some highlights of SemBERT are:
\begin{itemize}
    \item Encodes semantic role in an entity embedding that is combined at the output
\end{itemize}
\section{Combination and Hybrid Injections}\label{combinationinjections}
Here we describe approaches that use combinations of injection types such as input/output injections or architecture/output injections, etc. We start by looking at models that perform input injections and reinforce these with output injections (LIBERT, KALM),  We then look at models that manipulate the attention mechanisms to mimic graph connections (BERT-MK,K-BERT). We follow this by looking into KG-BERT, a model that operates on KG triples, and K-Adapter, a modification of RoBERTa that encodes KGs into Adapter Layers and fuses them. After this, we look into the approach presented as Cracking the Contextual Commonsense Code[...] which determines that there are areas lacking in BERT that could be addressed by supplying appropriate data, and we look at ERNIE 2.0, a framework for multi-task training for semantically aware models. Lastly, we look at two hybrid approaches which extract LM knowledge and leverage it for different tasks. A summary of these injections can be found in table \ref{tab:combinationinjections}.
\subsection{Knowledge-Aware Language Model (KALM) Pre-Training \cite{rosset2020knowledge}}
 KALM is a system that does not modify the internal architecture of the model that it is looking to inject the knowledge into, rather it modifies the input of the model by fusing entity embeddings with the normal word embeddings that the language model (in KALM's case, GPT-2) uses.  They then enforce the model in the output to uphold the entity information by adding an additional loss component in the pre-training that uses a max margin between the cosine distance of the output contextual representation and the input entity embedding and the cosine distance of the contextual representation and a confounder entity.  Altogether what this does is that it forces the model to notice when there is an entity and tries to make the contextual representation have the semantics of the correct input entity.  Some highlights of KALM are:
\begin{itemize}
    \item Sends an entity signal in the beginning and  and enforces  it in the output of a generative model to notice its semantics
\end{itemize}
\subsection{Exploiting  structured  knowledge  in  text  via  graph-guided   representation   learning\cite{shen2020exploiting}}
This work masks informative entities that are drawn from a knowledge graph, in BERT's MLM objective.  In addition to this, they have an auxiliary objective which uses a max-margin loss for a ranking task which is composed of using a bilinear model that calculates a similarity score between a contextual representation of an entity mention and the representation of the $[CLS]$ token for the text. The use for this is to determine if it is a relevant entity or a distractor.  Both KALM and this work are very similar, but a key difference is that KALM uses a generative model without any kind of MLM objective, and KALM does not do any kind of filtering for the entities. Some highlights of this work are:
\begin{itemize}
    \item Filters relevant entities to incorporate their information into the model
    \item Enforces entity signal at beginning and end of the model through masking and max-margin losses
\end{itemize}
\subsection{Lexically Informed BERT (LIBERT)\cite{lauscher2019informing}}
LIBERT converts batches of lexical constraints and negative examples, into a BERT-compatible format. The lexical constraints are synonyms and direct hyponyms-hypernyms (specific,broad) and take the form of a set of tuples of words: $(C = \{(w_1, w_2)_i\}^N_{
i=1})$. In addition to this set, the authors generate some negative examples by finding the words that are semantically close to $(w_1)$ and $(w_2)$ in a given batch.  They then format the examples into something BERT can use which is simply the wordpieces that pertain to words in the batch separated by the separator token. They pass this input through BERT and use the $[CLS]$ token as an input to a softmax classifier to determine if the example is a valid lexical relation or not.  

During pre-training they alternate between a batch of sentences and a batch of constraints.  LIBERT outperforms BERT with lesser (1M) iterations of pre-training.  It is worth noting that as the amount of iterations of training increase, the gap between the two systems, although present, becomes smaller.  This may indicate that, although the additional training objective is effective, it may be getting overshadowed by the regular MLM coupled with large amounts of data, however more testing needs to be performed. It is also worth noting that the authors do not align the sentences with the constraint batches, combine the training tuples which may hinder training as BERT has to alternate between different training input structures, and lastly, they do not incorporate antonymy constraints in their confounder selection, so further experimentation would be required to verify the effects of these. Some highlights of LIBERT are the following:
\begin{itemize}
    \item Incorporate lexical constraints from entity embeddings
    \item Good performance with constrained amounts of data
\end{itemize}

\subsection{BERT-MK\cite{he2019integrating}}
BERT-MK utilizes a combination of architecture injection and an output injection (additional training loss).  In BERT-MK, they utilize the KG-transformer modules which are transformer layers that are combined with learned entity representations. These entity representations are generated from another set of transformer layers that are trained on a KG converted to natural language sentences.  The interesting aspect is that these additional layers incorporate an attention mask that mimics the connections in the KG, so to a certain extent, incorporating the structure of the graph and propagating it back into the embeddings. These additional layers are trained to reconstruct the input set of triples. The authors evaluate the system for medical knowledge (MK) however it may be interesting to evaluate this on the GLUE benchmark along with utilizing other KGs such as ATOMIC or ConceptNet.  Some highlights of BERT-MK are:
\begin{itemize}
    \item Utilization of a modified attention mechanism to mimic KG structure between terms
    \item Incorporation of triple reconstruction loss to train the KG-transformer modules
    \item Merges KG-transformer with regular transformer for contextual+knowledge-informed representation
\end{itemize}

\begin{table*}[ht!]
\small
\hskip-2.0cm
    \begin{tabularx}{1.3\textwidth}{|m{1.75cm}|m{1cm}|m{1cm}|X|X|X|X|}
\hline
  \textbf{Model} & \textbf{Injected in Pre-Training}& \textbf{Injected in Fine-Tuning}& \textbf{Summary of Input Injection}& \textbf{Summary of Architecture Injection}& \textbf{Summary of Output Injection}\\\hline
  KALM&Yes&No&Combines an entity embedding with the model's word embeddings&N/a&Incorporates a max-margin loss with cosine distances to enforce semantic information\\\hline
   Exploiting structured knowledge in text [...]&Yes&No&Uses a KG informed masking scheme to exploit MLM learning&N/A&Incorporates a max-margin loss with distractors from a KG and a bilinear scoring model for the MM loss. \\\hline
   LIBERT&Yes&No&Alternates between batches of sentences, and batches of tuples that are lexically related&N/a&Adds a binary classifier as a third training task to determine if the tuples form a valid lexical relation\\\hline
   BERT-MK&Yes&No&N/A&Combines a base transformer with KG transformer modules which are trained to learn contextual entity representations and have an attention mechanism that mimics the connections between a graph&Use a triple reconstruction loss, similar to MLM, but for triples\\\hline K-BERT &Yes &No&Incorporate as part of their training batches, assertions from entities present in a sample&Modify the attention mechanism and position embeddings to reorder injected information, and mimic KG connections&N/A\\\hline KG-BERT &No&Yes&Feeds triples from a knowledge graph as input examples&N/A&Uses a binary classification objective to determine if a triple is correct and a multi-class classification objective to determine what kind of relation the triple has.\\\hline
   K-Adapter&No&Yes&N/A&Fine tune adapter layers to store information from a KG & Combine the model and different adapter layers to give a contextual representation with information from different sources, additionally use a relation classification loss for each trained adapter.\\\hline
   Cracking the contextual commonsense code[...]&Yes&No&Pre-process the data to address commonsense relation properties that are deficient in BERT &N/A&Concatenates a graph embedding to the output of the BERT model\\\hline ERNIE 2.0&Yes&No&Construct data for pre-training tasks &N/A&Provides a battery of tasks that are trained in parallel to enforce different semantic areas in a model\\\hline
   
  \end{tabularx}
    \caption{Combination Injection Systems Comparisons}
    \label{tab:combinationinjections}
\end{table*}

\subsection{K-BERT \cite{liu2020k}}
K-BERT uses a combination of input injection and architecture injections.  For a given sentence, they inject relevant triples for the entities that are present in the sentence and in a KG. They inject these triples in between the actual text and utilize a soft-position embedding to determine the order in which the triples are evaluated. These soft position embeddings simply add positional embeddings to the injected triple tokens.  This in turn creates a problem that the tokens are injected as entities appear in a sentence, and hence the ordering of the tokens is altered.  

To remedy this the authors utilize a masked self attention similar to BERT-MK. What this means is that the attention mechanism should only be able to see everything up to the entity that matched in the injected triple.  This attention mechanism helps the model focus on what relevant knowledge it should incorporate.  It would have been good to see a comparison of just adding these as sentences in the input rather than having to fix the attention mechanism to compensate for the erratic placement.  Some highlights of K-BERT are:
\begin{itemize}
    \item Utilization of attention mechanism to mimic connected subgraphs of injected triples
    \item Injection of relevant triples as text inputs
    
\end{itemize}
\subsection{KG-BERT \cite{yao2019kg} }
The authors present a combination approach which fine-tunes a BERT model with the text of triples from a KG similar to COMET. The authors also feed confounders in the form of random samples of entities into the training of the system. It utilizes a binary classification task to determine if the triple is valid and a relationship type prediction task to determine which relations are present between pairs of entities.  Although this system is useful for KG completion, there is no evidence of its performance on other tasks. Additionally they train one triple at a time which may limit the model's ability to learn the extended relationships for a given set of entities.   Some highlights of KG-BERT are the following:
\begin{itemize}
    \item Fine tunes BERT into completing triples from a KG
    \item Uses a binary classification to predict if a triple is valid
    \item Uses multi-class classification to predict relation type
\end{itemize}
\subsection{K-Adapter \cite{wang2020k}}
 A work based on adapters, K-Adapter works by adding projection layers before and after a subset of transformer layers.  They do this only for some specific layers in a pre-trained RoBERTa model (the first layer, the middle layer, and the last layer). They then freeze RoBERTa as per the Adapter work in \cite{houlsby2019parameter} and train 2 adapters to learn factual knowledge from Wikipedia triples \cite{elsahar2019t} and linguistic knowledge from outputs of the Stanford parser\cite{chen2014fast}. They then train the adapters with a triple classification (whether the triple is true or not) task similar to KG-BERT.

It is worth noting that the authors compare RoBERTa and their K-Adapter approach against BERT, and BERT has considerably better performance on the LAMA probes.  The authors attribute RoBERTa's byte pair encodings\cite{shibata1999byte} (BPE) as the major performance delta between their approach and BERT. Another possible reason may be that they only perform injection in a few layers rather than throughout the entire model, although testing needs to be done to confirm this. Some highlights of K-Adapter are:
\begin{itemize}
    \item Approach provides a framework for continual learning
    \item Use a fusion of trained adapter outputs for evaluation tasks 
\end{itemize}
\subsection{Cracking the Contextual Commonsense Code: Understanding Commonsense Reasoning Aptitude of Deep Contextual Representations\cite{da2019cracking}}
 The authors analyze BERT to determine that it is deficient in certain attribute representations of entities. The authors use the RACE \cite{lai2017race} dataset, and based on five attribute categories (Visual,  Encyclopedic, Functional Perceptual, Taxonomic), select samples from the dataset that may be help a BERT model compensate deficiencies in the areas. They then fine tune on this data. In addition to this, the authors concatenate the fine tuned BERT embeddings with some knowledge graph embeddings.  These graph embeddings are generated based on assertions that involve the entities that are present in the questions and passages they train their final joint model on (MCScript 2.0\cite{ostermann2019mcscript2}).  Their selection of additional fine-tuning data for BERT improves their performance in MCScript 2.0, highlighting that their selection addressed missing knowledge. 

It is worth noting that the graph embeddings that they concatenate boost the performance of their system which shows that there is still some information in KGs that is not in BERT.  We classify this approach as a combination approach because they concatenate the result of the BERT embeddings and KG embeddings and fine tune both at the same time. The authors however gave no insight as to how the KG embeddings could have been incorporated in the fine-tuning/pre-training of BERT with the RACE dataset.  Some highlights of this work are:
\begin{itemize}
    \item BERT has some commonsense information in some areas, but is lacking in others
    \item Fine-tuning on the deficient areas increases performance accordingly
    \item The combination of graph embeddings plus contextual representations are useful
\end{itemize}

\subsection{ERNIE 2.0 \cite{sun2020ernie}}
 The authors develop a framework that constructs pre-training tasks that center around Word-aware Pre-training, Structure-aware Pre-training, Semantic-aware Pre-training Tasks, and proceeds to train a transformer based model on these tasks.  An interesting aspect is that as they finish training on tasks, they keep training on older tasks in order for the model to not forget what it has learned.  In ERNIE 2.0 the authors do not incorporate KG information explicitly.  They do have a sub-task within the Word-aware pre-training that masks entities and phrases with the hope that it learns the dependencies of the masked elements which may help to incorporate assertion information.  

 A possible shortcoming of this model is that some tasks that are intended to infuse semantic information into the model (i.e. the Semantic aware tasks which are a Discourse Relation task and an information retrieval (IR) relevance) rely on the model to pick it up from the distributional examples. This could have the same possible issue as with the Input Injections and would need to be investigated further. Additionally, they do not explicitly use KGs in the work. Some highlights of ERNIE 2.0 are:
\begin{itemize}
    \item Continual learning platform keeps training on older tasks to maintain their information
    \item Framework permits flexibility on the underlying model
    \item Wide variety of semantic pre-training tasks 
\end{itemize}

\subsection{Graph-based reasoning  over  heterogeneous  external  knowledge for  commonsense  question  answering \cite{lv2020graph}}\label{subsec:graphbasedreasoning} 
A hybrid approach in which the authors do not inject knowledge into a language model (namely XLNet\cite{yang2019xlnet}, rather they utilize a language model as a way to unify graph knowledge and contextual information. They combine XLNet embeddings as nodes in a Graph Convolutional Network (GCN) to answer questions.  

They generate relevant subgraphs of ConceptNet and Wikipedia (from ConceptNet the relations that include entities in a question/answer exercise and  the top 10 most relevant sentences from ElasticSearch on Wikipedia).  They then perform a topological sorting on the combined graphs and pass them as input to XLNet.   XLNet then generates contextual representations that are then used as representations for nodes in a Graph Convolutional Network (GCN).  They then utilize graph attention to generate a graph level representation and combine it with XLNet's input ([CLS] token) representation to determine if an answer is valid for a question.  In this model they do not fine-tune XLNet, which could have been done on the dataset to give better contextual representations, and additionally they do not leverage the different levels of representation present in XLNet. 
Some highlights of this work are the following
\begin{itemize}
    \item Combination of GCN, Generative Language Model, and Search systems to answer questions
    \item Use XLNet as contextual embedding for GCN nodes
    \item Perform QA reasoning with the GCN output
\end{itemize}

\subsection{Commonsense knowledge base completion with structural and semantic context\cite{malaviya2020commonsense}}\label{subsec:cbkcwithstructure}
Another hybrid approach, the authors fine tune a BERT model on a list of the unique phrases that are used to represent nodes in a KG.  They then take the embeddings from BERT and from a sub-graph in the form of a GCN and run it through an encoder/decoder structure to determine the validity of an assertion.\footnote{It is worth noting that the two hybrid projects possibly benefited from the ability for these language models to encode assertions as shown by Feldman et al. \cite{davison2019commonsense} and Petroni et al.\cite{petroni2019language}.}

They then take this input and concatenate it with node representations for for a subgraph (in this case a combination of ConceptNet and Atomic).  They treat this concatenation as an encoded representation, and run combinations of these through a convolutional decoder that additionally takes an embedding of a relation type. 

The result of the convolutional decoder is run through a bilinear model and a sigmoid  function to determine the validity of the assertion.  It seems interesting that the authors only run the convolution through one side: convolution of $(e_i,e_{rel})$ rather than both the convolution of $(e_i,e_{rel})$ and $(e_{rel},e_j)$ (where $(e_i,e_j)$ are the entity embeddings for entity i and j respectively and $(e_{rel})$ is the embedding for a specific relationship) and then a concatenation.  They rely on the bilinear model joining the two representations.

Some highlights of this work are the following:
\begin{itemize}
    \item Use a GCN and a LM to generate contextualized assertions representations
    \item Use BERT to generate contextual embeddings for nodes
    \item Use an encoder-decoder structure to learn triples
\end{itemize}

\section{Future Directions}
\subsection{Input Injections}
Most input injections are to format KG information into whatever format a transformer model can ingest.  Although KALM has explored incorporating a signal to the input representations, it would be interesting to add additional information such as the lexical constraints mentioned in LIBERT, to the word embeddings that are trained with the transformer based models like BERT.  A possible approach could be to build a post-specialization system that could generate retrofitted\cite{faruqui2014retrofitting} representations that can then be fed into language models.  

\subsection{Architecture Injections}
Adapters seem to be a promising field of research in language models overall.  The idea that one can fine tune a small amount of parameters may simplify the injection of knowledge and KnowBERT has explored some of these benefits. It would be interesting to apply a similar approach to generative models and see the results.

Another possible avenue of research would be to incorporate neural memory  models/modules such as the ones by Munkhdalai\cite{munkhdalai2019metalearned} into adapter-based injections. The reasoning would be that the model can simply look up relevant information encoded into a memory architecture and fuse it into a contextual representation. 

\subsection{Combined Approaches} There are a variety of combined approaches, but none of them tackle all three areas (input, architecture, and output) at the same time.  It seems promising to test out a signaling method such as KALM and see how this would work with an adapter based method similar to KnowBERT.  The idea being that the input signal could help the entity embeddings contextualize better within the injected layers.  Additionally, it would be interesting to see how the aforementioned combination would look with a system similar to LIBERT; such that you could fuse entity embeddings with some semantic information.  
\section{Conclusion}
Infusing structured information from Knowledge Graphs into pre-trained language models has had some success.  Overall, the works reviewed here give evidence that the models benefit from the incorporation of the structured information. By analyzing the existing works, we give some research avenues that may help to develop more tightly coupled language/KG models.  
\bibliography{bibliography}
\clearpage
\appendix
\begin{table}[h!]
\small\hskip-2.0cm
    \begin{tabularx}{1.3\textwidth}{|X|X|X|X|X|}
\hline
  \textbf{Knowledge Injection Approach} & \textbf{Underlying Language Model}& \textbf{Type of Injection}&\textbf{Knowledge Sources}&\textbf{Training Objective}\\
\hline
Align, Mask, Select	& BERT &	Input & ConceptNet &	Binary Cross Entropy\\\hline
COMET&	GPT&	Input&	Atomic, ConceptNet & Language Modeling\\\hline
KnowBERT&	BERT&	Architecture&	Wikipedia, WordNet & Masked Language Modeling\\\hline 
Common  sense  or  world  knowledge?[...] & BERT&	Architecture&ConceptNet	&Masked Language Modeling\\\hline
SemBERT& BERT& Output& Semantic Role Labeling of Pre-Training data & Masked Language Modeling\\\hline
KALM&	GPT-2&	Combination (Input,Output)&	 FACC1 and FAKBA entity annotation\cite{gabrilovich2013facc1}& Language Modeling + Max Margin\\\hline
Exploiting Structured[...]&	BERT&	Combination (Input+Output)	& ConceptNet& Masked Language Modeling, Max Margin \\\hline
LiBERT&	BERT&	Combination (Input, Output)&Wordnet, Roget's Thesaurus&	Masked Language Modeling + Max Margin\\\hline
Graph-based reasoning  over  heterogeneous  external  knowledge &	XLNet + Graph Convolutional Network&	Hybrid (Language Model + Graph Reasoning)&	Wikipedia, ConceptNet &Cross Entropy\\\hline

Commonsense knowledge base completion with structural and semantic context& BERT+Graph Convolutional Net&	Hybrid (Language Model + GCN Embeddings)& Atomic,ConceptNet	&Binary Cross Entropy\\\hline

ERNIE 2.0&	Transformer-Based Model & Combination (Input+Output)&	Wikipedia, BookCorpus, Reddit, Discovery Data (Various types of relationships extracted from these datasets) &Various tasks, among them Knowledge Masking, Token-Document Relation Prediction, Sentence Distance Task, IR Relevance Task\\\hline

BERT-MK&	BERT&	Combination (Architecture+Output)	&Unified Medical Language System\cite{bodenreider2004unified}(UMLS) & Masked Language Modeling, Max Margin \\\hline

K-Bert&	BERT&	Combination (Input + Architecture)&TBD	&Same As BERT\\\hline

KG-Bert&	BERT&	Combination (Input + Output)&Freebase, Wordnet, UMLS 	&Binary + Categorical Cross Entropy\\\hline
K-Adapter&	RoBERTa&	Combination (Architecture+Output)& Wikipedia, Dependency Parsing from Book Corpus&	Binary Cross Entropy\\\hline

Cracking the Commonsense Code&	BERT&	Combination (Input+Output)&N/A:Fine Tuning on RACE dataset subset&	Binary Cross Entropy\\\hline

\end{tabularx}
\caption{Knowledge Injection Models Overview}
\end{table}

\begin{table}[h!]
\small
\hskip-2.0cm
    \begin{tabularx}{1.3\textwidth}{|X|X|X|X|X|X|}
\hline
\textbf{Knowledge Injection Approach}& \textbf{Benchmark Name} & \textbf{Base Model }&\textbf{Base Model Benchmark Performance}& \textbf{Knowledge Injected Model Performance}&\textbf{Percent Difference}\\\hline

BERT-CS$_{base}$ 

(Align, Mask, Select)&GLUE (Average)&Bert-Base&78.975&79.612&0.81\% \\\hline

BERT-CS$_{large}$

(Align, Mask, Select)&GLUE (Average)&Bert-Base-large&81.5&81.45&-0.06\% \\\hline

LIBERT (2M) &GLUE (Average)&BERT Baseline Trained with 2m examples&72.775&74.275&2.06\%\\\hline
SemBERT$_{base}$&GLUE(Average)&BERT-Base&78.975&80.35&1.74\%\\\hline
SemBERT$_{large}$&GLUE(Average)&BERT-Base&81.5&84.262&3.39\%\\\hline
K-Adapter F+L&CosmosQA, TACRED&RoBERTa + Multitask training &81.19,71.62&81.83,71.93&1.54\%,	0.95\%\\\hline
Ernie 2.0 (large)&GLUE (Average) & BERT-Base-Large & 81.5&84.65&3.87\%\\\hline
BERT-MK&Entity Typing, Rel. Classification (using UMLS)&BERT-base&96.55	,77.75&97.26,83.02&0.74\%,6.78\%\\\hline
K-BERT&XNLI&BERT-base&75.4&76.1&0.93\%\\\hline
Cracking the contextual commonsense code (Bert large + KB+RACE)&MCScript 2.0&BERT-Large&82.3&85.5&3.89\\\hline
KnowBERT&TACRED&BERT-Base&66&71.5&8\%\\\hline
Common Sense or World Knowledge? (OM-ADAPT 100K)&GLUE (Average)&BERT-Base&78.975&79.225&0.40\%
\\\hline
Common Sense or World Knowledge? (CN-ADAPT 50K)&GLUE (Average)&BERT-Base&78.975&79.225&0.32\%\\\hline
\end{tabularx}
\caption{Knowledge Injection Models Performance Comparison}
\end{table}

\end{document}